%% file: main.tex
\documentclass{article}
\usepackage{PRIMEarxiv}

\usepackage{amsmath,amssymb,amsfonts}
\usepackage{graphicx}
\usepackage{textcomp}
\usepackage{xcolor}
\def\BibTeX{{\rm B\kern-.05em{\sc i\kern-.025em b}\kern-.08em
    T\kern-.1667em\lower.7ex\hbox{E}\kern-.125emX}}

\input{packages.tex}

\usepackage{subcaption} 
\usepackage{minted}
\usepackage{tabularx}
\usepackage{makecell}
\usepackage{array}
\newcolumntype{P}[1]{>{\raggedright\arraybackslash}p{#1}}

\usepackage{hyperref}

\begin{document}

\title{Search-based Robustness Testing of Laptop \\ Refurbishing Robotic Software}


\author{
  Erblin Isaku \\
  Simula Research Laboratory and \\
  University of Oslo \\
  Oslo, Norway \\
  \texttt{erblin@simula.no}
  \And
  Hassan Sartaj \\
  Simula Research Laboratory \\
  Oslo, Norway \\
  \texttt{hassan@simula.no}
  \And
  Shaukat Ali \\
  Simula Research Laboratory \\
  Oslo, Norway \\
  \texttt{shaukat@simula.no}
  \And
  Malaika Din Hashmi \\
  Danish Technological Institute \\
  Odense, Denmark \\
  \texttt{mdih@teknologisk.dk}
  \And
  Francois Picard \\
  Danish Technological Institute \\
  Odense, Denmark \\
  \texttt{fpi@teknologisk.dk}
  \And
}



\maketitle
\newcommand{\approach}{{\fontfamily{qhv}\selectfont PROBE}}

\begin{abstract}
The Danish Technological Institute (DTI) focuses on transferring advanced technologies (including robots) to the industry and the public sector. One key application is laptop refurbishment using specialized robots, aimed at promoting reuse, reducing electronic waste, and supporting the European Circular Economy Action Plan. The software of such robots often includes features that use object detection models to detect objects for various purposes, such as identifying screws for laptop disassembly or detecting stickers to remove them. Ensuring the robustness of such models to small input variations remains a critical challenge, and addressing it is important to avoid potential damage to laptops during refurbishment. In this paper, we propose \approach{}, a search-based robustness testing approach that leverages multi-objective optimization to identify minimal, localized perturbations that expose failures in object detection models used in the software of laptop refurbishing robots. \approach{} employs NSGA-II to systematically explore the perturbation space, optimizing for failure induction considering both localization and confidence, and perturbation magnitude, while enabling the discovery of diverse failure cases. 
Results show that \approach{} is 3× to 7× more effective than random search in generating failure-inducing perturbations, while requiring smaller perturbation magnitudes, and that the generated perturbations transfer across models. 
We further show that metamorphic relations provide additional insights into model robustness, enabling the assessment of stability even in non-failing cases.
\end{abstract}





\section{Introduction}

The Danish Technological Institute (DTI) is a Danish institute providing industry-ready solutions across several areas, including artificial intelligence and robotics, for a wide range of social and industrial applications. To this end, DTI is also a leader in developing robotic applications for sustainability, such as reducing electronic waste and refurbishing equipment to enable reuse, including laptops, which is the application area addressed in this paper. The laptop refurbishing robots are designed to work collaboratively with humans to automate the refurbishment of older laptops for resale, including disassembly, cleaning (including removing stickers), and reassembly. 

Software in laptop refurbishing robots plays a key role in automating various steps. A key step in refurbishing is identifying objects, such as screws, so that a robot can unscrew them and remove components, such as screens, and detecting stickers attached to laptops so they can be removed. Such software includes object detection models used to identify objects on laptops. Ensuring the successful identification of such objects is important, since incorrect detection can cause robots to damage laptops, for example, when removing stickers from a laptop’s surface~\cite{lu2025assessing, lu2026uamters}. To this end, it is important to ensure that such models are robust against minor variations in their inputs to avoid incorrect object detection. Such robustness can be ensured through automated and systematic testing of these models, which is the focus of this paper.

In current DTI practice, robustness is addressed with basic data augmentation techniques (e.g., flipping, exposure adjustment, or blurring) applied during training to enrich datasets and improve generalization. These techniques are applied in an ad hoc and unsystematic manner, without a structured, automated way to assess robustness or identify failures. While these techniques are valuable for simulating variability in the absence of large-scale real-world data~\cite{tanaka2019data, maharana2022review}, they are not explicitly designed for testing. As a result, they provide limited support for systematically uncovering failure-inducing conditions, particularly subtle ones that may arise in practice (e.g., varying lightning conditions or surface wear).
Motivated by the need to improve current practice, 
we propose \approach{}, a multi-objective search-based approach for robustness testing of DL-based object detection software in robots.
\approach{} systematically explores the input space by generating structured, minimal perturbations aligned with industrial practice to uncover failure-inducing cases. In addition, it enables a comprehensive robustness analysis by characterizing different failure types (i.e., misclassification, mislocalization, missed detection, and ambiguous predictions) and by assessing the stability of confidence and localization scores under non-failing perturbations. \looseness=-1

We evaluated \approach{} using three industrial object detection models, i.e., \textit{origSCDM}, \textit{noscrewSCDM}, and \textit{fixtureSCDM}, provided by DTI. 
Results show that \approach{} consistently generated a larger number of failure-inducing cases, achieving failure rates between 10.4\% and 26.3\% compared to 1.5\% to 9,9\% for the baseline, i.e., random search, while requiring smaller perturbation magnitudes. Furthermore, the generated perturbations exhibit stronger transferability across models, indicating their applicability and effectiveness across different object detection models used in DTI. 
Beyond failure cases, \approach{} reveals distinct failure behaviors across models. For \textit{origSCDM}, failures are primarily driven by misclassifications ($\approx 7.2$ per image) and ambiguous predictions ($\approx 2.0$ per image). A similar pattern is observed for \textit{noscrewSCDM}, with misclassifications ($\approx 5.7$ per image) and ambiguous predictions ($\approx 2.6$ per image). In contrast, failures in \textit{fixtureSCDM} are dominated by localization errors ($\approx 2.7$ per image). 
Finally, results on non-failing perturbations indicate that \textit{fixtureSCDM} is the most robust model, exhibiting very high stability in both confidence and localization outputs, with only around 1\% confidence violations and near-zero localization violations. Based on our results, we also provide lessons learned that we consider beneficial for both researchers and practitioners. 

\section{Industrial Context}


This work is conducted in collaboration with the Danish Technological Institute (DTI), a research and innovation organization focused on transferring advanced technologies to industry, particularly in robotics for sustainable and circular operations. Within the European Circular Economy Action Plan (CEAP), improving the refurbishment of electronic equipment is a critical challenge, as it is one of the fastest-growing waste streams in the EU, with less than 40\% recycled~\cite{ref_ceap}. Circular economy initiatives are projected to increase the EU’s GDP by 0.5\% and create approximately 700,000 jobs by 2030~\cite{ref_ceap}.
In this context, laptop refurbishment represents a relevant and challenging application domain, where DTI is developing an automated system based on a UR10e collaborative robot equipped with vision sensors (Intel RealSense) to support tasks such as screen replacement across different laptop models. 

\begin{figure}
  \centering
    \includegraphics[width=0.65\columnwidth]{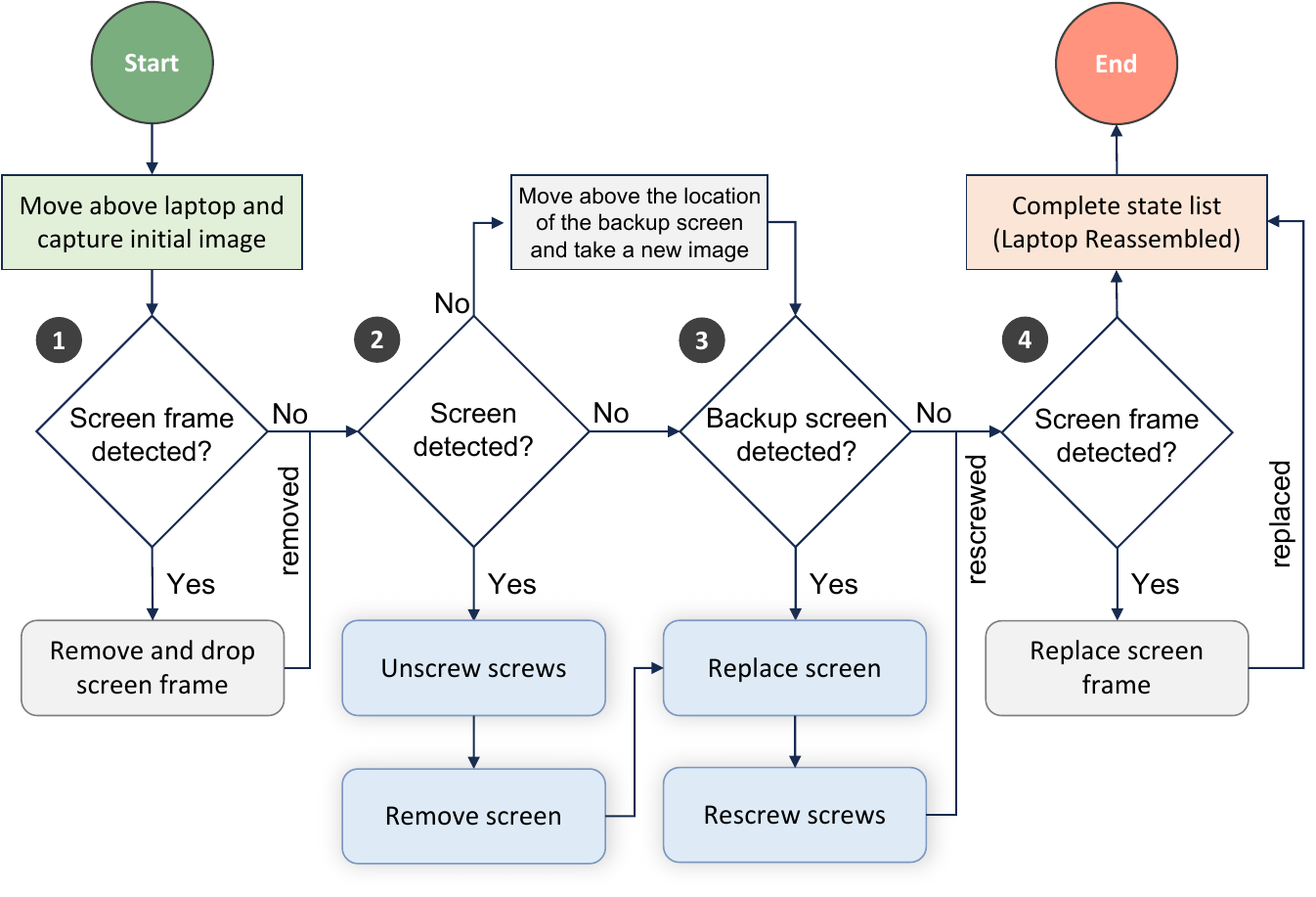}
    \caption{Laptop disassembly process workflow of the refurbishment robotic system from DTI.
    \label{fig:dti-workflow}}
\end{figure}

The overall disassembly workflow is illustrated in Figure~\ref{fig:dti-workflow}, which captures the sequence of perception-driven decisions and manipulation steps required to remove and replace the laptop screen. The process begins by capturing an initial image, followed by a series of decision points (Steps 1–4) based on visual detections, including \textit{screen frame}, \textit{screen}, and \textit{backup screen}. Depending on these outcomes, the system executes corresponding actions such as removing the screen frame, unscrewing screws, removing the screen, replacing it with a backup, and finally reassembling the components (e.g., rescrewing screws and replacing the frame) until reaching the final reassembled state.

In this work, we specifically focus on the perception component corresponding to screw detection, which supports the blue-highlighted operations within the workflow. As shown in the figure, this component directly supports critical operations such as \textit{unscrew screws} and \textit{rescrew screws}, making it a key enabler of the overall disassembly process.

A key component of the disassembly workflow is AI-based object detection software that relies on sensor data. However, it faces several challenges in the real-world refurbishment process:
\textit{(a) sensitivity to environmental conditions}, where variations in lighting and sensor noise can significantly affect detection performance; \textit{(b) high variability across devices}, as used laptops differ in structure, wear, and prior repairs, making generalization difficult; \textit{(c) failure propagation}, where errors in screw detection directly impact downstream operations (e.g., unscrewing and reassembly), potentially leading to incomplete execution, increased human intervention, or damage to components; and \textit{(d) lack of systematic robustness assessment}, which limits the ability to identify and address failure modes under representative, domain-specific operating conditions.


To address these challenges, this work focuses on the screw detection component, based on YOLO object detection models~\cite{redmon2016you}, and assesses its robustness by identifying minimal perturbations that induce failures or degrade performance.
Identified perturbations can also reveal previously unseen failure modes and may be incorporated into the training pipeline to improve model robustness. By systematically uncovering such weaknesses, our approach enables a better understanding of the system’s failure boundaries and supports the development of more reliable and adaptable robotic solutions for real-world refurbishment scenarios.

\section{Approach}
\subsection{Approach Overview}
\label{sec:approach_overview}

\begin{figure*}[t]
  \centering
    \includegraphics[width=1.0\textwidth]{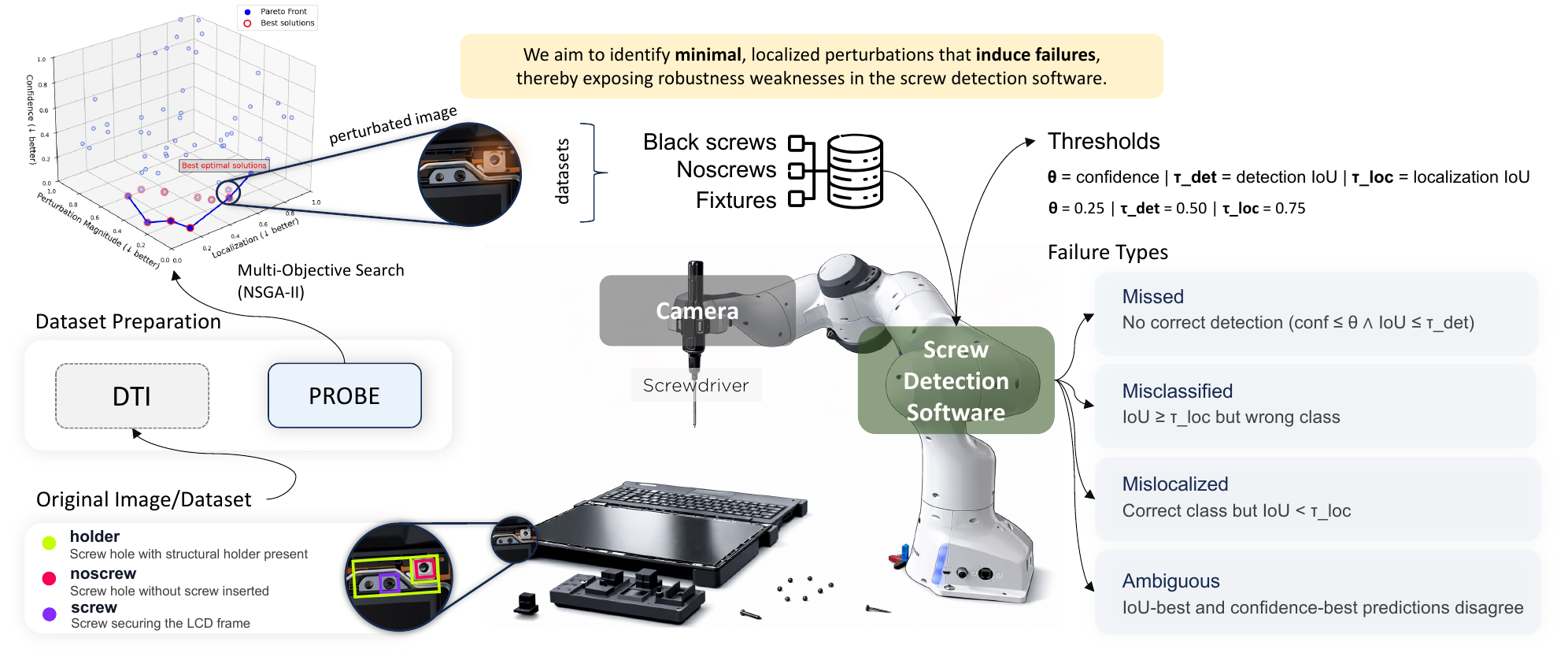}
    \caption{Overview of \approach{}, a search-based robustness testing approach for the screw detection component in the laptop refurbishment software.}
    \label{fig:approach-overview}
\end{figure*}

Figure~\ref{fig:approach-overview} presents \approach{}'s overview 
Starting from an original image and its annotations (i.e., holder, noscrew, screw), \approach{} applies a multi-objective search to identify minimal and localized perturbations that expose weaknesses in the screw detection software.
Focusing on minimal changes allows us to uncover subtle, failure-inducing conditions that are more likely to occur in real-world settings (e.g., variations in lightning), rather than trivial failures caused by overly strong perturbations.
The goal is not simply to generate arbitrary corrupted inputs, but to systematically uncover perturbations that induce failures or degrade detection performance.

At a high level, \approach{} operates in three stages. First, it takes as input an image from the industrial dataset along with its corresponding ground-truth annotations. Second, it uses NSGA-II to explore a space of localized perturbations and evaluate them according to multiple objectives that capture both their impact on model performance and their perturbation magnitude. In particular, the search aims to reduce detection confidence and localization quality while keeping the perturbation small and spatially constrained, as these objectives balance the ability to induce failure with the need to keep perturbations minimal and localized.  \looseness=-1

Finally, the resulting perturbed inputs are analyzed using predefined thresholds to determine whether they lead to failures such as missed detections (i.e., failure to detect an existing object), misclassifications (i.e., incorrect class prediction), mislocalizations (i.e., inaccurate bounding box), or ambiguous cases (i.e., conflicting predictions for the same object). \looseness=-1

As a result, \approach{} supports both failure discovery and robustness analysis. On the one hand, it identifies perturbations that can fail the perception component under small input variations. On the other hand, it provides insight into how close the robotic software is to failure by exposing cases where minimal perturbations already degrade confidence or localization. In this way, \approach{} offers a systematic way to characterize robustness weaknesses, including failures, in the perception pipeline used in the laptop refurbishment workflow (Figure~\ref{fig:dti-workflow}).

\subsection{Problem Formulation}
\label{sec:problem_formulation}

Let $x \in \mathcal{X}$ denote an input image and $y$ its corresponding ground-truth annotations, consisting of a set of labeled bounding boxes. Let $M$ be a perception model that, given an input image, produces a set of predictions $\hat{y} = M(x)$, where each prediction includes a bounding box, a class label, and a confidence score. We assess the robustness of $M$ under small, localized input perturbations by defining a parameterized perturbation $p \in \mathcal{P}$ that transforms the original image $x$ into a perturbed image $x' = T(x, p)$, where $T$ is a perturbation function. The perturbation space $\mathcal{P}$ is bounded to ensure that modifications are limited in magnitude and spatial extent.

Given an image $x$ and its annotations $y$, we formulate robustness testing as a multi-objective optimization problem over the perturbation space $\mathcal{P}$. Specifically, we seek to identify perturbations $p$ that degrade the model’s performance while remaining minimal and localized. This is achieved by minimizing the following objective vector:
\begin{equation}
\min_{p \in \mathcal{P}} \; \mathbf{f}(p) = \left(f_1(p), f_2(p), f_3(p)\right)
\label{eq:optimization}
\end{equation}

where:

\begin{itemize}
    \item $f_1(p)$ measures the average confidence of correct (same-class) detections,
    \item $f_2(p)$ measures the average localization quality in terms of Intersection over Union (IoU),
    \item $f_3(p)$ quantifies the perturbation budget, capturing both its magnitude and spatial extent.
\end{itemize}

These objectives are chosen to capture complementary aspects of object detection performance and perturbation characteristics.
In particular, $f_1$ and $f_2$ capture the degradation of the model’s predictions under perturbation, while the third objective constrains the perturbation to remain small and localized. By jointly optimizing these objectives, the approach identifies perturbations that expose weaknesses in the model without introducing large or global modifications to the original image.\looseness=-1

The search is restricted to a region of interest defined by the ground-truth bounding boxes, a common practice where domain knowledge guides the search towards relevant regions to reduce computational cost~\cite{harman2009theoretical}. In our context, this constraint reduces the search space size and directs the perturbation towards areas that directly influence the detection task. This makes the search more efficient, and the identified perturbations are easier to relate to the model’s predictions and potential failure cases.

The solution to this multi-objective problem is a set of Pareto-optimal perturbations, each representing a different balance between performance degradation and perturbation magnitude. These solutions are then analyzed to identify failure cases and characterize the robustness of the perception component.

\subsection{Solution Encoding}
\label{sec:encoding}

Each candidate solution $p \in \mathcal{P}$ represents a localized perturbation applied to the input image. We encode $p$ as a seven-dimensional vector:
\begin{equation}
p = (c_x, c_y, r, \sigma, \Delta B, \Delta G, \Delta R)
\label{eq:perturbation-vector}
\end{equation}
where $(c_x, c_y)$ denotes the center of the perturbation, $r$ is its radius, $\sigma$ controls the spatial spread, and $\Delta B$, $\Delta G$, and $\Delta R$ represent additive color perturbations applied to the blue, green, and red channels, respectively.

Given a candidate $p$, the perturbation function $T(x,p)$ generates a perturbed image $x'$ by applying a spatially localized modification centered on $(c_x, c_y)$. Specifically, a Gaussian mask is constructed to smoothly distribute the perturbation within a circular region of radius $r$. The standard deviation of the Gaussian is defined as a fraction of the radius, i.e., $\sigma = \alpha r$, where $\alpha$ is a bounded ratio. This results in a smooth transition between perturbed and unperturbed regions.
The perturbation is applied by adding a color offset to each pixel, weighted by the Gaussian mask. Formally, for each pixel location $(i,j)$:
\begin{equation}
x'(i,j) = \mathrm{clip}\left(x(i,j) + \alpha(i,j)\,{\delta}, \; 0, \; 255\right),
\label{eq:perturbation}
\end{equation}
where ${\delta} = [\Delta B, \Delta G, \Delta R]$ denotes the channel-wise perturbation applied to the blue, green, and red components, and $\alpha(i,j)$ is the Gaussian mask value at the pixel location $(i,j)$. The operation is applied element-wise to each color channel. The function $\mathrm{clip}(\cdot, 0, 255)$ restricts the resulting pixel values to the valid image range by truncating values outside the interval $[0, 255]$. 

To ensure that perturbations remain small and localized, each parameter is constrained within predefined bounds (Table~\ref{tab:perturbation-params}). The perturbation center $(c_x, c_y)$ is restricted to lie within a region of interest (ROI), defined by the ground-truth bounding boxes and expanded by a fixed margin. The remaining parameters are bounded to control the spatial extent, smoothness, and magnitude of the perturbation.

\begin{table}[htbp]
\centering
\renewcommand{\arraystretch}{1.2}
\caption{Perturbation parameter bounds used in \approach{}.}
\label{tab:perturbation-params}
\begin{tabular}{l l l}
\hline
\textbf{Parameter} & \textbf{Range} & \textbf{Description} \\
\hline
$c_x, c_y$ & ROI $\pm$ margin (5 px) & Perturbation center \\
$r$ & $[8, 80]$ & Patch radius \\
$\alpha$ & $[0.15, 0.80]$ & Gaussian spread ratio \\
$\Delta B, \Delta G, \Delta R$ & $[-48, 48]$ & Channel-wise intensity shift \\
\hline
\end{tabular}
\end{table}
The parameter bounds are selected based on domain knowledge, aligned with current industrial practice, and informed by discussions with industrial partners. In particular, the region-of-interest margin accounts for annotation imprecision, while the radius and spread parameters control the spatial extent and smoothness of the perturbation. The color perturbation range is bounded relative to the image intensity scale to ensure moderate modifications. These choices balance the need to effectively influence model predictions while avoiding overly large or global perturbations.
It is important to note that these bounds are not tuned for specific images, but are fixed across all experiments and methods to ensure consistency and fairness. \looseness=-1

\subsection{Objective Functions}\label{ssec:objfunc}


Given a perturbed image $x' = T(x,p)$, the perception model $M$ produces a set of predictions $\hat{y} = M(x')$. Each prediction consists of a bounding box, a class label, and a confidence score. Let $y$ denote the set of ground-truth objects in the image.

\paragraph{Confidence Objective ($f_1$).}
The first objective measures the average confidence of correct (same-class) detections. For each ground-truth object, we identify the predicted bounding box of the same class that maximizes the Intersection over Union (IoU). The confidence score associated with this best match is recorded and set to zero if no such prediction exists. The objective is defined as:
\begin{equation}
f_1(p) = \frac{1}{|y|} \sum_{i=1}^{|y|} \mathrm{conf}_i,
\label{eq:f1}
\end{equation}
where $\mathrm{conf}_i$ denotes the confidence of the best same-class prediction for object $i$.

\paragraph{Localization Objective ($f_2$).}
The second objective measures the localization quality of the detections. For each ground-truth object, we compute the IoU of the best same-class prediction. If no such prediction exists, the IoU is set to zero. The objective is defined as:
\begin{equation}
f_2(p) = \frac{1}{|y|} \sum_{i=1}^{|y|} \mathrm{IoU}_i,
\quad \text{where } 
\mathrm{IoU}(A, B) = \frac{|A \cap B|}{|A \cup B|}.
\label{eq:f2}
\end{equation}
where $\mathrm{IoU}_i$ is the intersection over union between the ground-truth bounding box and the corresponding best same-class predicted bounding box for object $i$, with $A$ and $B$ denote the ground-truth and predicted bounding boxes, respectively.

\paragraph{Perturbation Budget ($f_3$).}
The third objective quantifies the magnitude and spatial extent of the perturbation. The spatial extent is measured as the fraction of pixels affected by the perturbation, capturing how much of the image is modified. The perturbation intensity is estimated using the $95^{\text{th}}$ percentile of the absolute perturbation magnitude within the affected region, providing a robust measure of strong but representative changes while reducing sensitivity to outliers.
The intensity term is normalized by the maximum allowed perturbation magnitude $\epsilon$, which corresponds to the upper bound of the color perturbation range (i.e., $\Delta c \in [-\epsilon, \epsilon]$ in Table~\ref{tab:perturbation-params}). This normalization expresses the perturbation strength relative to the defined search space, ensuring comparability across candidate solutions. The objective is defined as:
\begin{equation}
f_3(p) = \text{area\_frac}(p) \cdot \frac{\text{mag}_{95}(p)}{\epsilon},
\label{eq:f3}
\end{equation}
where $\text{area\_frac}(p)$ denotes the fraction of pixels influenced by the perturbation, computed as the ratio between the number of affected pixels and the total number of pixels in the image. The term $\text{mag}_{95}(p)$ represents the $95^{\text{th}}$ percentile of the absolute perturbation magnitude within the affected region.

All objectives in Eqs.~\eqref{eq:f1}, \eqref{eq:f2}, and \eqref{eq:f3} are minimized during the search. The first two objectives capture the degradation of detection performance in terms of confidence and localization, respectively, while the third objective constrains the perturbation to remain small and spatially limited. By jointly optimizing these objectives, the search process identifies perturbations that expose weaknesses in the model while maintaining controlled modifications to the input.

\section{Experiment Design}
\label{sec:experiment_design}

\subsection{Research Questions}

We evaluate the effectiveness of \approach{} in uncovering robustness issues (i.e., failures and performance degradation) in object detection models. The goal of \approach{} is to identify minimal, localized perturbations that induce failures, and to analyze their characteristics and impact on model behavior. Accordingly, we define and address the following research questions (RQs).

\vspace{0.5em}

\begin{itemize}
    \item[\textbf{RQ1}] \emph{How effective is \approach{} compared to random search in discovering failure-inducing perturbations?}
    \item[\textbf{RQ2}] \emph{How effectively does \approach{} discover different types of failures?}
    \item[\textbf{RQ3}] \emph{To what extent are metamorphic relations violated under minimal, non-failing perturbations?}
\end{itemize}

RQ1 investigates whether the search space of perturbations requires a multi-objective optimization approach (\approach{}), or whether a simpler baseline, random search, is sufficient to discover failure-inducing perturbations. To this end, we evaluate the approach along three dimensions:
(a) \textit{search quality}, measured using Hypervolume (HV), which captures the ability of each method to explore high-quality trade-offs across objectives;
(b) \textit{effectiveness}, measured in terms of (i) failure rate, indicating how frequently failure-inducing perturbations are discovered, and (ii) perturbation magnitude, reflecting how subtle or minimal such perturbations are. 
(c) \textit{transferability}, assessed through cross-model experiments (e.g., perturbations generated for model $M_X$ and applied to model $M_Y$).
\looseness=-1

RQ2 examines the ability of \approach{} to uncover and characterize diverse failure types in object detection models, including missed detections, misclassifications, mislocalizations, and ambiguous cases.

RQ3 focuses on whether minimal perturbations that do not induce failures still lead to violations of expected model behavior. 
Since not all robustness issues manifest as explicit failures, and acceptable deviations in model outputs are often not well defined (i.e., lacking an oracle), we employ metamorphic relations (MRs) to capture subtle inconsistencies in model behavior.
For example, \textit{a screw may still be correctly detected under a small perturbation, but with a significantly shifted bounding box, indicating degraded robustness without an explicit failure.}
To this end, we evaluate two MRs: (a) \textit{confidence stability}, which requires that prediction confidence remains within a threshold $\epsilon_{\text{conf}}$; and (b) \textit{localization stability}, which requires that bounding box localization remains consistent within an IoU threshold $\epsilon_{\text{iou}}$.
This analysis provides insights into the stability of the models, even in cases where predictions remain formally correct.

\subsection{Experimental Setup}

\subsubsection{Use Cases and Datasets}


To assess the robustness of the proposed approach, we consider three YOLOv11 object detection models from the industrial partner (DTI), trained on progressively enriched datasets. These models capture different levels of variability in the training data, ranging from simple scenarios with only screws to more complex settings including non-screw objects and diverse transformations (e.g., screw fixtures).

Table~\ref{tab:datasets} summarizes the characteristics of the datasets used in the evaluation.
For each model, a subset of images is selected from the corresponding training dataset by retaining a single representative version per source image, avoiding redundancy introduced by data augmentation. This results in a combined evaluation set of 802 images across all models, enabling a comprehensive robustness analysis.
In addition, a separate test dataset consisting of 25 images from the industrial case study is used to evaluate cross-model transferability \textit{RQ1c}. This dataset is disjoint from the training data of all models, ensuring a fair assessment of generalization.
All datasets and models are publicly available and can be accessed online~\cite{ref_dataset_v0, ref_dataset_v1, ref_dataset_v2a, ref_dataset_v2b}.\looseness=-1









\begin{table}[htbp]
\centering
\caption{Overview of the object detection models and dataset characteristics used in the evaluation. Data augmentations include Flip (Horizontal, Vertical), Exposure ($\pm$10\%), Gaussian blur (0--2.5 px), Crop (0--20\%), and Rotation ($\pm 15^\circ$ and $90^\circ$).}
\label{tab:datasets}
\renewcommand{\arraystretch}{1.1}
\setlength{\tabcolsep}{4pt}
\small
\begin{tabularx}{\textwidth}{
>{\raggedright\arraybackslash}X
r
r
>{\raggedright\arraybackslash}X}
\hline
\textbf{Dataset (Description)} & \textbf{Training} & \textbf{Eval.} & \textbf{Augmentations} \\
\hline
Black screws only (\textit{origSCDM})~\cite{ref_dataset_v0}
& 989 & 331
& Flip, Exposure \\

Screws + non-screw (\textit{noscrewSCDM})~\cite{ref_dataset_v1}
& 1189 & 400
& Flip, Blur \\

Extended with fixtures (\textit{fixtureSCDM})~\cite{ref_dataset_v2a, ref_dataset_v2b}
& 419 & 71
& Flip, Crop, Rotation, Blur \\
\hline
\end{tabularx}

\vspace{2pt}
\footnotesize
\noindent\textit{Note:} The training-derived dataset contains 802 images and is used in \textit{RQ1a}, \textit{RQ1b}, \textit{RQ2}, and \textit{RQ3}. The test dataset contains 25 images and is used in \textit{RQ1c}.
\end{table}

\subsubsection{Search Algorithms}

We implement the proposed approach (\approach{}) using the \texttt{pymoo} optimization framework~\cite{pymoo}, which includes NSGA-II for multi-objective optimization.
\approach{} maintains a population of candidate perturbations and iteratively evolves them using non-dominated sorting and crowding distance to approximate a diverse set of Pareto-optimal solutions. New candidate solutions are generated using standard evolutionary operators, including crossover and mutation, as defined by the framework.

As a baseline, we consider Random Search (RS), which samples perturbations uniformly at random from the defined search space. In contrast to \approach{}, RS does not utilize any feedback from previous evaluations and therefore does not guide the search towards optimal regions.

\subsubsection{Parameter Settings}

For a fair comparison between search strategies, we executed \approach{} and RS with the same computational budget and configurations.
\approach{} was configured with a population size of 40 and evolved for 500 generations (determined through a pilot study to ensure convergence), resulting in a total of 20,000 fitness evaluations per run. Random Search is executed with the same number of evaluations to ensure a fair comparison. 

For the initial population, each candidate solution $p \in \mathcal{P}$ is sampled uniformly at random within the predefined parameter bounds (Table~\ref{tab:perturbation-params}) to ensure that all initial solutions satisfy the constraints on spatial location, size, smoothness, and perturbation magnitude.
In particular, the perturbation center $(c_x, c_y)$ is sampled within the region of interest defined by the ground-truth bounding boxes, while the remaining parameters are sampled within their respective ranges. This initialization strategy provides a diverse set of candidate perturbations, allowing the search process to explore different regions of the solution space from the outset. All experiments were repeated for 10 independent runs to account for randomness in the search process~\cite{arcuri2011practical}. The default configuration of the \texttt{pymoo} framework was used for the evolutionary operators in NSGA-II.

\subsubsection{Execution Details}
Each execution of \approach{} produces a set of Pareto-optimal perturbations per image. Given a population size of 40 and 10 independent runs, this results in up to 400 candidate solutions per image. Across the full evaluation set of 827 images, this corresponds to approximately 330{,}800 perturbed images generated by NSGA-II.
RS was executed in the same configuration, producing an equivalent number of perturbed images. In total, both search algorithms generate approximately 661{,}600 perturbed images, which are used to evaluate the robustness of the screw detection system.

\subsection{Evaluation Metrics}


\paragraph{Quality Indicator}

\textit{Hypervolume (HV)} is used to evaluate the quality of the approximated Pareto front in the multi-objective setting~\cite{shang2020survey}. It measures the volume of the objective space dominated by the obtained solutions with respect to a reference point. In our case, a higher HV value indicates better coverage of trade-offs between reducing confidence, degrading localization, and limiting perturbation magnitude.

\paragraph{Model Failure and Stability}

\textit{Failure Rate} measures the proportion of perturbations that result in incorrect detection outcomes. A failure is defined as any deviation from the expected prediction, including missed detections, misclassifications, incorrect localization, or ambiguous predictions. Formally, the failure rate for a model $M$ (i.e., \textit{origSCDM, noscrewSCDM, fixtureSCDM}) is defined as
\begin{equation}
FR_M = \frac{\sum_{i=1}^{N} y_i}{N},
\end{equation}
where $y_i \in \{0,1\}$ indicates whether perturbation candidate $i$ induces a failure, and $N$ is the total number of generated perturbation candidates across all images (and runs) for model $M$. \looseness=-1

In addition, we analyze \textit{failure occurrences} to capture how many object-level failures are induced by each perturbation. Let $f_i^{(t)}$ denote the number of failures of type $t$ (i.e., \textit{missed}, \textit{mislocalized}, \textit{misclassified}, or \textit{ambiguous}) caused by perturbation candidate $i$. 
The definition of each failure type is illustrated in Figure~\ref{fig:approach-overview}.
The total number of failure occurrences of type $t$ for a model $M$ is defined as
\begin{equation}
FO_M^{(t)} = \sum_{i=1}^{N} f_i^{(t)},
\end{equation}
where $N$ is the total number of perturbation candidates across all images (and runs) for model $M$.
While the failure rate reflects whether a perturbation leads to a robotic software failure or not, failure occurrences quantify the number and type of failures and are used to analyze the distribution of failure types in RQ2.


\paragraph{Confidence and Localization Stability}

To assess robustness under non-failing perturbations, we measure the deviation between predictions on original and perturbed images. Each perturbation candidate $i$ corresponds to a perturbed version of an input image. Let $c_i$ and $\tilde{c}_i$ denote the confidence scores before and after perturbation, respectively, and let $\text{IoU}_i$ and $\widetilde{\text{IoU}}_i$ denote the corresponding localization quality. The confidence and localization deviations are defined as:
\begin{equation}
\Delta_{\text{conf}}(i) = | c_i - \tilde{c}_i | ;\quad
\Delta_{\text{loc}}(i) = | \text{IoU}_i - \widetilde{\text{IoU}}_i |
\end{equation}

These deviations are used to categorize perturbations into different stability levels (minor, small, moderate, and large), based on predefined thresholds.

\paragraph{Statistical Analysis}

To assess whether the observed differences between approaches are statistically significant, we employ non-parametric statistical tests.
For paired comparisons (e.g., Hypervolume and failure rate computed per image and run), we use the \textit{Wilcoxon signed-rank test}. For unpaired comparisons (e.g., perturbation magnitude computed over independent sets of failure-inducing perturbations), we use the \textit{Mann--Whitney U test}.
In addition to statistical significance, we report effect sizes to quantify the magnitude of the differences. For paired comparisons, we use the \textit{rank-biserial correlation ($r_{rb}$)}, computed from the normalized difference between the number of wins and losses. For unpaired comparisons, we use the \textit{Vargha--Delaney effect size ($\hat{A}_{12}$)}. Following standard guidelines, we interpret effect sizes as negligible, small, medium, or large~\cite{arcuri2011practical}.

\section{Results and Analyses}

\subsection{RQ1 - Comparison with Random Search}
\input{rq1-summary}

\subsubsection{Search Quality (HV)}

Table~\ref{tab:rq1_summary} shows that \approach{} consistently outperforms RS across all models in terms of HV. For \textit{origSCDM}, \approach{} achieves a mean HV of 0.1474 compared to 0.0935 for RS, while for \textit{noscrewSCDM} the values are 0.1119 and 0.0559, respectively. For \textit{fixtureSCDM}, although the absolute difference is smaller (0.0249 vs. 0.0197), \approach{} still achieves consistently higher HV values. These differences are statistically significant in all cases (Wilcoxon signed-rank test, $p < 0.001$). Moreover, the large effect sizes (rank-biserial correlation ranging from 0.8510 to 0.8967) indicate that \approach{} outperforms RS in the majority of paired comparisons, demonstrating not only improved performance but also strong consistency across runs and images.


\subsubsection{Effectiveness}

We analyze failure rates to assess each approach's ability to discover failure-inducing perturbations. Table~\ref{tab:rq1_summary} shows a consistent and statistically significant advantage of \approach{} over RS. For \textit{origSCDM}, \approach{} achieves a mean failure rate of 26.3\% compared to 9.9\% for RS, while for \textit{noscrewSCDM} the corresponding values are 23.3\% and 6.4\%. For \textit{fixtureSCDM}, failure rates are lower overall, with 10.4\% for \approach{} and 1.5\% for RS, suggesting that this is the most robust model among the three.
These differences are statistically significant in all cases (Wilcoxon signed-rank test, $p < 0.001$), with large effect sizes ($r_{rb} \in [0.8957, 0.9560]$), indicating that \approach{} discovers failures more frequently in all paired comparisons. Furthermore, when considering only failure-inducing perturbations, \approach{} consistently produces smaller perturbation magnitudes than RS, with statistically significant differences (Mann--Whitney U test, $p < 0.001$). This confirms that \approach{} not only finds more failures but also identifies more minimal and thus more subtle perturbations.





\subsubsection{Transferability Across Models}

\input{rq1_transferability}

To evaluate whether failure-inducing perturbations generalize across models, we generate perturbations on a source model and evaluate them on different target models (Table~\ref{tab:rq1_transferability}). Overall, \approach{} consistently achieves higher transferability than RS across all source--target pairs.

The diagonal entries show that \approach{} is more effective even when perturbations are evaluated on the same model on which they were generated, achieving higher failure rates (42.9\%, 37.3\%, and 15.2\% for \textit{origSCDM}, \textit{noscrewSCDM}, and \textit{fixtureSCDM}, respectively) compared to RS (24.3\%, 11.8\%, and 0.7\%). This advantage extends to cross-model transfer. For instance, perturbations generated on \textit{noscrewSCDM} transfer to \textit{origSCDM} with a failure rate of 29.8\% (vs.\ 16.0\% for RS), while those generated on \textit{origSCDM} transfer to \textit{noscrewSCDM} with 16.6\% (vs.\ 6.9\%). Across all off-diagonal entries, \approach{} consistently outperforms RS, indicating that it identifies perturbations that generalize across models.

Transferability is, however, asymmetric and depends strongly on the target model. In particular, \textit{fixtureSCDM} is the most robust target, with very low transfer rates (0.8\% and 2.4\%) despite higher rates in the reverse direction. This is consistent with its lower failure rate when evaluated on its own perturbations (15.2\%), compared to 42.9\% for \textit{origSCDM} and 37.3\% for \textit{noscrewSCDM}. In contrast, \textit{origSCDM} is the most vulnerable target, showing higher transfer rates from other models. \looseness=-1


\begin{tcolorbox}[colframe=black!50, colback=gray!5, boxrule=0.3mm]
\approach{} more effectively explores the search space, uncovers more failures with smaller perturbation magnitude, and produces perturbations that transfer better across models compared to Random Search.
\end{tcolorbox}

\subsection{RQ2 - Failure Discovery}

\begin{figure*}[tbp]
  \centering
    \includegraphics[width=1\textwidth]{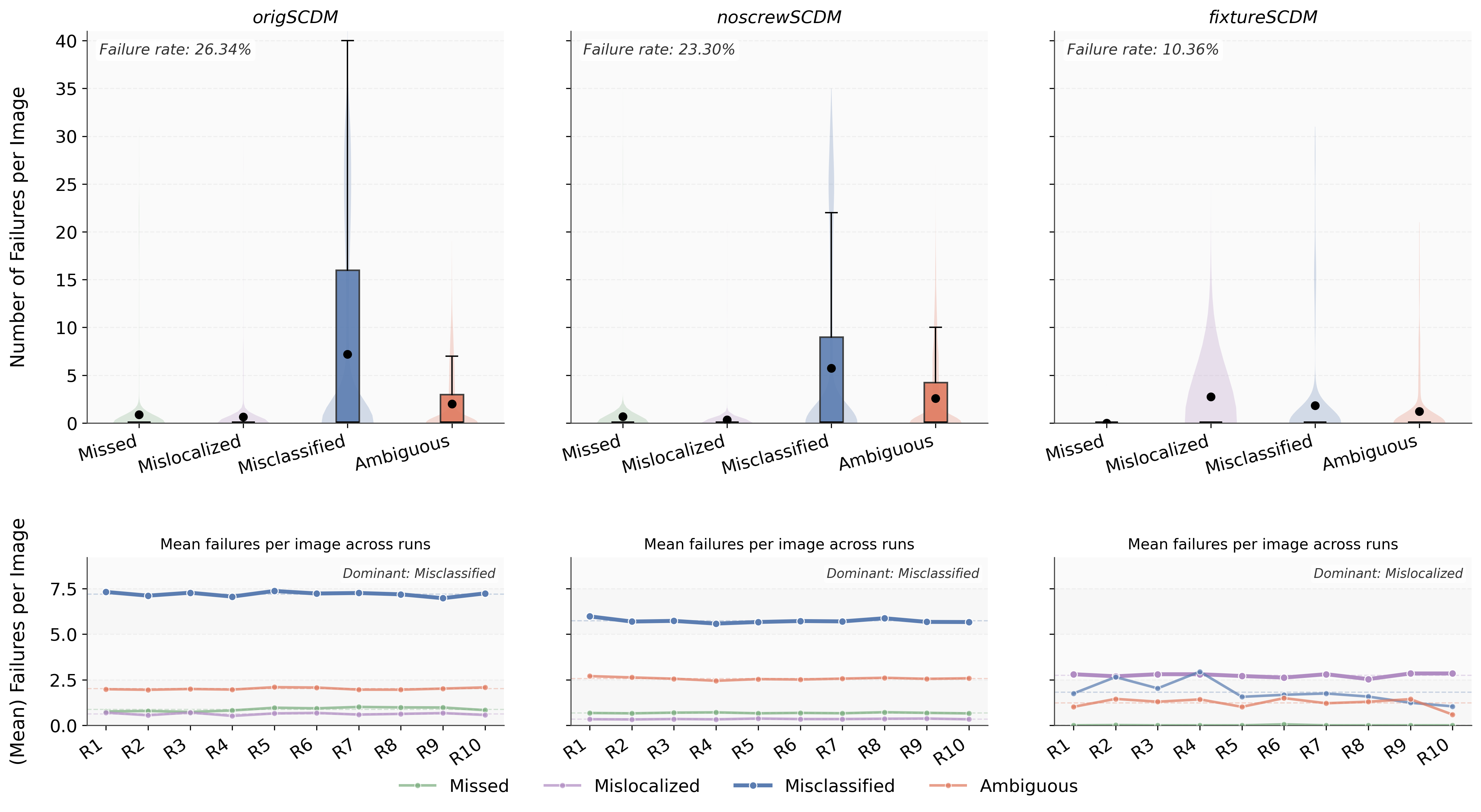}
    \caption{
    Distribution and consistency of failure types across models. The top row shows the per-image distribution of failure types (Missed, Mislocalized, Misclassified, Ambiguous) aggregated over all runs, using violin and box plots to illustrate variability across images (40 perturbation candidates per image). The annotated failure rate indicates the overall proportion of failing perturbations. The bottom row presents the mean number of failures per image across runs (Run 1–10), highlighting the consistency and stability of failure patterns across runs. Dominant failure types are indicated for each model (RQ2).
}
    \label{fig:rq2_failure_types}
\end{figure*}

To better understand the nature of the failures discovered by \approach{}, we analyze the distribution of failure types across runs and models (Figure~\ref{fig:rq2_failure_types}) in four categories: \textit{missed}, \textit{mislocalized}, \textit{misclassified}, and \textit{ambiguous} detections.

Overall, the results show that \approach{} consistently discovers failures across all categories. For \textit{origSCDM}, the overall failure rate is $\approx$ 26.3\% (34{,}880 failing perturbations), while the total number of failure occurrences is slightly higher (35{,}464), indicating that some perturbations trigger multiple failures. The majority of failures are due to misclassifications (mean $\approx 7.2$ per image), followed by ambiguous cases ($\approx 2.0$), while missed and mislocalized detections occur much less frequently (both below $1.0$ on average). A similar pattern is observed for \textit{noscrewSCDM}, with an overall failure rate of 23.3\% (37{,}288 failing perturbations), where misclassifications again dominate (mean $\approx 5.7$), followed by ambiguous cases ($\approx 2.6$).

In contrast, \textit{fixtureSCDM} exhibits a substantially lower overall failure rate of $\approx$ 10.4\% (2{,}941 failing perturbations and 4{,}114 failure occurrences), confirming its higher robustness. Moreover, the distribution of failure types is more balanced, with lower absolute counts across all categories. In this case, mislocalized detections dominate (mean $\approx 2.7$ per image), while misclassifications occur less frequently (mean $\approx 1.8$). Ambiguous cases remain relatively limited (mean $\approx 1.2$), and missed detections are negligible. These results indicate that, although failures are less frequent overall, they are primarily driven by localization deviations rather than classification mistakes. \looseness=-1
Across all models, the failure distributions are highly consistent across runs, with only minor variations for \textit{fixtureSCDM}. This indicates that \approach{} produces stable and reliable results in terms of the number and types of failures.

\begin{figure}[t]
\vspace{0.3em}
  \centering
    \includegraphics[width=0.65\textwidth]{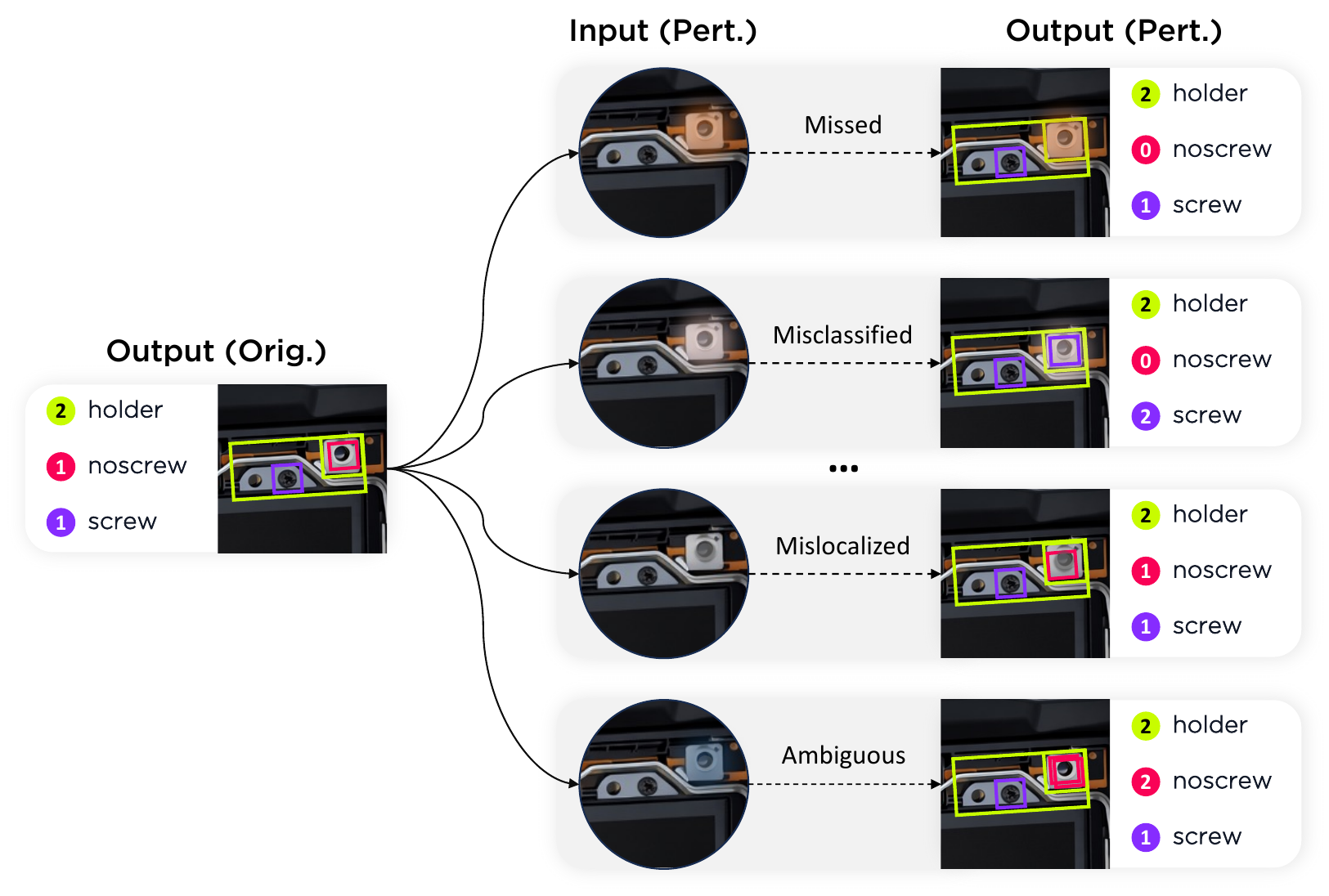}
    \caption{Illustration of the failure types identified by \approach{}. Each example shows the original input and prediction, followed by the perturbed input and resulting prediction. Note that the images are adapted for presentation purposes; the original data can be accessed via the datasets listed in Table~\ref{tab:datasets}.}
    \label{fig:failure-types}
\end{figure}

To complement the quantitative results, Figure~\ref{fig:failure-types} illustrates representative examples of the four failure types identified by \approach{}. Starting from the prediction on the original (unperturbed) image, where all objects are correctly detected, we show how perturbations lead to different failure types. In \textit{missed} detections, an object (e.g., \textit{noscrew}) is no longer identified despite remaining visible. In \textit{misclassified} cases, the object is still localized but assigned an incorrect label. \textit{Mislocalized} detections correspond to correct classification with a shifted bounding box, as shown for the \textit{noscrew} instance where the predicted box is displaced, reducing overlap with the ground truth. Finally, \textit{ambiguous} detections arise when multiple conflicting predictions are produced for the same region, such as duplicate \textit{noscrew} detections with differing confidence and IoU. \looseness=-1
Overall, these examples highlight how perturbations can affect both classification and localization, and although each case is shown independently, a single perturbed input may impact multiple objects within an image, leading to different failure behaviors across instances.

\begin{tcolorbox}[colframe=black!50, colback=gray!5, boxrule=0.3mm]
Failures discovered by \approach{} occur across all categories, with model-specific patterns. Misclassifications dominate for \textit{origSCDM} and \textit{noscrewSCDM}, while \textit{fixtureSCDM} is mainly affected by mislocalizations. Overall, \textit{fixtureSCDM} shows higher robustness.
Additionally, some perturbations affect multiple objects within the same image.
\end{tcolorbox}


\subsection{RQ3 - Robustness Analysis via Metamorphic Relations}

\subsubsection{Violation of Metamorphic Relations}

In metamorphic testing, a violation of an MR is typically defined as a deviation from the expected behavior under a given transformation. However, in practice, not all deviations necessarily indicate a violation. Small variations may occur without affecting the overall correctness of the system’s output.
For this reason, we interpret MR violations in terms of \textit{stability thresholds}. Instead of treating every deviation as a violation, we categorize changes into different levels (minor, small, moderate, and large, as shown in~\cref{tab:rq3_stability}) based on predefined thresholds. 
Importantly, the choice of threshold is context-dependent and should be determined by domain experts based on acceptable levels of variation. In safety-critical settings, stricter thresholds may be required, whereas in other contexts, moderate deviations may still be acceptable.
In this work, we consider an (MR) violation when the observed deviation falls into the \textit{moderate} or \textit{large} category, for either confidence or localization stability. Minor and small deviations are treated as acceptable variations that do not constitute a violation.

\input{rq3_confidence}

\subsubsection{Confidence and Localization Stability}


Overall, most perturbations result in minor or small deviations, indicating stable model behavior under non-failing conditions. In particular, minor deviations (e.g., $\leq 0.01$) correspond to very small changes in confidence or bounding box position, while small deviations (0.01–0.05] reflect limited variation that does not significantly affect predictions. 

Focusing on MR violations (moderate and large deviations), \textit{origSCDM} exhibits 17.8\% confidence violations and 10.5\% localization violations. Similarly, \textit{noscrewSCDM} shows 9.8\% and 17.5\%, respectively. These cases indicate situations where predictions remain correct but show noticeable degradation, such as a significant drop in confidence or a visible shift in the bounding box. For example, a detected screw may have IoU = 0.92 in the original image, which drops to 0.78 after perturbation (a deviation of 0.14), indicating a significant localization shift despite correct detection. 
Such deviations may affect the quality of the process in practice, potentially leading to incomplete operations or requiring human intervention.
In contrast, \textit{fixtureSCDM} remains highly stable, with only $\approx$ 1\% confidence violations and near-zero localization violations (0.02\%). These results indicate that violations are relatively limited overall, with \textit{fixtureSCDM} demonstrating the highest robustness, particularly in localization stability.

\begin{tcolorbox}[colframe=black!50, colback=gray!5, boxrule=0.3mm]
Models remain largely stable under non-failing perturbations, with most confidence and localization deviations falling within minor and small ranges. Metamorphic violations (moderate and large deviations) are limited, with \textit{origSCDM} and \textit{noscrewSCDM} showing a higher number of violations.
\end{tcolorbox}

\section{Threats to Validity}

\textit{Internal validity} threats may arise from the implementation of \approach{} and the selection of perturbation and search parameters. To mitigate this, we used consistent experimental settings, validated results over multiple runs to reduce randomness, and conducted a pilot study to determine appropriate search parameters (e.g., number of generations) based on the convergence behavior of the search objectives. 
\textit{External validity} is limited by the use of three object detection models from a single industrial case study (DTI), and results may not generalize to other domains, datasets, or model architectures. Future work should investigate the generalizability of our approach across broader contexts (e.g., vision-language-action models). 
\textit{Construct validity} concerns the choice of metrics used to assess robustness, including failure rate, perturbation magnitude, and stability measures ($\epsilon_{\text{conf}}$ and $\epsilon_{\text{iou}}$). While the selected metrics are motivated by domain-specific considerations and industrial relevance, they may not capture all aspects of robustness in object detection systems. 
Finally, \textit{conclusion validity} is supported by the use of statistical tests and effect size measures, as well as consistent trends observed across models, although results may still be influenced by dataset characteristics and experimental settings.

\section{Discussion and Lessons Learned}
\paragraph*{\bf Lesson 1: Perturbations can induce multiple concurrent failures}
Our analysis revealed that individual perturbations can induce multiple concurrent failures, as evidenced by the total number of failure occurrences exceeding the number of failing perturbations in \textit{origSCDM} and \textit{fixtureSCDM}. 
This observation suggests that most perturbations affect only part of the image, despite the presence of multiple objects (e.g., holder and screw/no-screw). Practitioners shall explore complementing single-object perturbations with multi-object, region-aware perturbations that simultaneously target all relevant scenes, thereby exposing more complex failure cases. \looseness=-1

\paragraph*{\bf Lesson 2: Failure-inducing perturbations can generalize across models} Perturbations generated by \approach{} can be transferred across different models; however, we find that such transferability is not symmetric across models, i.e., perturbations may transfer well from one model to another but not vice versa. Our results suggest that practitioners can reuse perturbations generated for one model version in subsequent versions, reducing the need to regenerate them for each new version and thereby improving testing efficiency. However, this does not necessarily guarantee that the reused perturbations ensure the robustness of the newer model; instead, they can serve as a useful starting point for robustness assessment.     


\paragraph*{\bf Lesson 3: Laptop refurbishment robots operate in highly uncertain and unstructured environments, which makes ensuring robustness of perception challenging} Laptop refurbishment robots must handle laptops with significant variability in models, wear conditions, and physical configurations (e.g., screws), as well as user-specific appearances (e.g., attached stickers). This variability can substantially degrade perception performance in practice, as indicated by our results, highlighting the need for systematic and automated robustness evaluation methods such as the one proposed in this paper.

\paragraph*{\bf Lesson 4: Bridging testing and data augmentation practices}
\approach{} advances current industrial practice by shifting perturbations from training-only data augmentation (Table~\ref{tab:datasets}) to systematic testing. While existing augmentations improve robustness during training, they are not used to evaluate model behavior under controlled variations. \approach{} enables targeted discovery of failure-inducing cases and produces perturbations that can be reused to strengthen robotic software (e.g., by integrating failure cases into subsequent training iterations). This suggests that practitioners should adopt methods like \approach{} as part of their CI/CD process, where discovered failures are fed back into training to continuously improve model robustness. \looseness=-1

\section{Related Work}

Search-based software testing (SBST) treats testing problems as optimization problems and solves them using search algorithms to efficiently explore large input spaces~\cite{harman2012search}.
In this context, SBST has been applied to core testing activities such as test generation~\cite{ali2009systematic}, selection and prioritization~\cite{lu2021search}, and fault localization~\cite{liu2017improving}, enabling efficient and targeted discovery of faults. More recently, these techniques have been extended to deep learning-enabled systems, where they support the exploration of complex and high-dimensional input spaces~\cite{pei2017deepxplore, guo2018dlfuzz}. In particular, search-based approaches have been used to identify failure-inducing behaviors in deep learning models and autonomous systems, including applications in autonomous driving and robotics~\cite{tang2023survey, humeniuk2023ambiegen, lu2021search}, where robustness under varying conditions is critical~\cite{tian2018deeptest, zolfagharian2023search}. 
Closely related to our work, Wang et al.~\cite{wang2021robot} propose RobOT, a robustness-oriented testing approach that generates inputs via metric-guided mutations and integrates them into retraining pipelines. In contrast, our approach does not rely on access to model internals, but instead systematically explores the input space through structured, minimal perturbations, making it suitable for settings with limited model access. More broadly, prior work, including RobOT, has focused mainly on classification tasks and misprediction-oriented testing, whereas the application of search-based robustness testing to object detection systems remains underexplored--particularly in industrial workflows where such models operate as part of larger pipelines.

From a case study perspective, our prior work explored the laptop refurbishment scenario using object detection for tasks such as sticker detection and uncertainty quantification~\cite{lu2025assessing, lu2026uamters}. While these studies focused on prediction reliability and mutation-based evaluation, this work shifts toward systematic search-based robustness testing for failure discovery. In addition, we consider screw detection, which introduces greater complexity due to smaller object sizes and higher visual variability. By enabling targeted failure discovery and structured robustness analysis, this work extends our previous efforts toward more comprehensive validation of deep learning-enabled robotic systems in industrial settings.

\section{Conclusion and Future Work}
In this work, we presented \approach{}, a multi-objective search-based approach for robustness testing of object detection models in refurbishing robotic software at DTI. 
While \approach{} demonstrates promising results, several directions for future work remain:
\begin{inparaenum}
    \item extend \approach{} to other perception components and model architectures, e.g., sticker detection models based on Faster R-CNN~\cite{ren2015faster} architecture;
    \item integrate \approach{} into continuous testing pipelines at DTI, where discovered failures are included in training to improve model robustness; and
    \item explore self-adaptive methods that enable robotic software not only to detect but also to recover from such failures during operation.
\end{inparaenum}

\section*{Data Availability}
The datasets and object detection models used in this study are available at \cite{ref_dataset_v0, ref_dataset_v1, ref_dataset_v2a, ref_dataset_v2b}. The implementation of \approach{}, including the analyses scripts, is publicly available at \url{https://github.com/Simula-COMPLEX/PROBE}.

\section*{Acknowledgments}
This work is supported by the RoboSAPIENS project funded by the European Commission’s Horizon Europe programme under grant agreement number 101133807. Erblin Isaku is also supported by the Norwegian Ministry of Education and Research. 

\bibliographystyle{ACM-Reference-Format}
\bibliography{refs}

\end{document}

%% file: packages.tex
\usepackage[T1]{fontenc}
\usepackage[utf8]{inputenc}
\usepackage[american]{babel}
\usepackage{csquotes}
\usepackage{microtype}
\usepackage[defblank]{paralist}
\setdefaultenum{(1)}{(a)}{(i)}{A.}
\usepackage[inline]{enumitem}
\usepackage{url}
\usepackage{flushend}
\usepackage{etoolbox}
\usepackage{fix-cm}
\usepackage{textpos}
\usepackage{mfirstuc}
\usepackage{titlecaps}
\usepackage[datesep=.,style=ddmmyyyy]{datetime2}

\usepackage{algorithm}
\usepackage{algpseudocode}
\usepackage{pgf}

\usepackage{tcolorbox}

\newtcolorbox{Summary}{
    sharpish corners, 
    boxrule = 0pt, 
    toprule = 3.5pt, 
    toptitle = 1mm, 
    enhanced,
    fuzzy shadow = {0pt}{-2pt}{-0.5pt}{0.5pt}{black!35}, 
    colback = white, 
    colframe = gray, 
    coltitle=black, 
    fonttitle=\bfseries 
}
\usepackage[framemethod=tikz]{mdframed}

\mdfdefinestyle{mpdframe}{
    frametitlebackgroundcolor   =black!15,
    frametitlerule              =true,
    roundcorner                 =5pt,
    middlelinewidth             =1pt,
    innermargin                 =0.3cm,
    outermargin                 =0.3cm,
    innerleftmargin             =0.3cm,
    innerrightmargin            =0.3cm,
    innertopmargin              =0.5cm,
    innerbottommargin           =0.5cm
}

\tcbuselibrary{skins}
\usetikzlibrary{shadings}

\colorlet{colexam}{red!75!black}

\usepackage{amsmath}
\usepackage{amssymb}
\usepackage{bm}
\usepackage{relsize}
\usepackage{siunitx}
\usepackage[super]{nth}

\usepackage{booktabs}
\usepackage{multirow}
\usepackage{colortbl}
\usepackage{makecell}
\usepackage{rotating}
\usepackage{threeparttable}
\usepackage{adjustbox}

\usepackage{float}
\usepackage{graphicx}
\usepackage{xcolor}

\usepackage{listingsutf8}

\usepackage{xparse}
\usepackage{xspace}

\usepackage[xindy,acronym]{glossaries}
\glsdisablehyper

\usepackage[numbers,sort&compress]{natbib}

\usepackage[hidelinks,bookmarks=false]{hyperref}

\usepackage[capitalise]{cleveref}

\definecolor{mycustomcolor}{HTML}{6D8764}

\tcbset{
  base/.style={
    enhanced,
    colframe=teal!5, 
    colback=teal!5,  
    sharp corners,
    borderline west={2pt}{0pt}{teal!70},
    boxrule=0.5mm,
    boxsep=4pt,
    leftrule=2mm,
    titlerule=0pt,
    fonttitle=\bfseries\itshape, 
    width=\textwidth, 
    attach boxed title to top left={
      xshift=2mm,
      yshift=-2mm
    }, 
    boxed title style={
      colback=teal!90, 
      colframe=teal!90, 
      sharp corners
    }
  }
}

\newtcolorbox{myexamplec}[2][]{
  base,
  title={#2}, 
}

%% file: rq1-summary.tex
\begin{table}[t]
\centering
\caption{Comparison of \approach{} and Random Search (RS) across models in terms of search quality (HV), failure rate, and perturbation magnitude. For HV and failure rate, effect size is reported as rank-biserial correlation ($r_{rb}$). For perturbation magnitude, effect size is reported as Vargha--Delaney $\hat{A}_{12}$.
$\uparrow$ indicates that higher scores are better, while $\downarrow$ indicates that lower scores are better.
$\triangledown$ indicates a small effect size, $\vartriangle$ indicates a medium effect size, and $\blacktriangle$ a large effect size (RQ1).}
\label{tab:rq1_summary}

\footnotesize
\renewcommand{\arraystretch}{1.1}
\setlength{\tabcolsep}{2pt}

\resizebox{\textwidth}{!}{%
\begin{tabular}{lcccccccccccc}
\toprule
& \multicolumn{4}{c}{\textbf{Hypervolume (HV) $\uparrow$}} 
& \multicolumn{4}{c}{\textbf{Failure Rate (\%) $\uparrow$}} 
& \multicolumn{4}{c}{\textbf{Perturbation Magnitude $\downarrow$}} \\
\cmidrule(lr){2-5} \cmidrule(lr){6-9} \cmidrule(lr){10-13}
\textbf{Model}
& \textbf{PROBE} & \textbf{RS} & \textbf{p-value} & \textbf{Effect Size}
& \textbf{PROBE} & \textbf{RS} & \textbf{p-value} & \textbf{Effect Size}
& \textbf{PROBE} & \textbf{RS} & \textbf{p-value} & \textbf{Effect Size} \\
\midrule

origSCDM
& \textbf{0.1474} & 0.0935 & $< 0.001$ & 0.8967 $\blacktriangle$
& \textbf{26.3} & 9.9 & $< 0.001$ & 0.8985 $\blacktriangle$
& \textbf{0.0112} & 0.0152 & $< 0.001$ & 0.6712 $\vartriangle$ \\

noscrewSCDM
& \textbf{0.1119} & 0.0559 & $< 0.001$ & 0.8510 $\blacktriangle$
& \textbf{23.3} & 6.4 & $< 0.001$ & 0.8957 $\blacktriangle$
& \textbf{0.0135} & 0.0218 & $< 0.001$ & 0.7018 $\vartriangle$ \\

fixtureSCDM
& \textbf{0.0249} & 0.0197 & $< 0.001$ & 0.8845 $\blacktriangle$
& \textbf{10.4} & 1.5 & $< 0.001$ & 0.9560 $\blacktriangle$
& \textbf{0.0039} & 0.0152 & $< 0.001$ & 0.5861 $\triangledown$ \\
\bottomrule
\end{tabular}%
}
\end{table}

%% file: rq1_transferability.tex

\begin{table}[ht]
\centering
\caption{Transferability of failure-inducing perturbations across models. Each cell reports the failure rate (\%) when perturbations generated on the source model (columns) are evaluated on the target model (rows). Results are aggregated over all images and runs (RQ1).}
\label{tab:rq1_transferability}
\renewcommand{\arraystretch}{1.1}
\setlength{\tabcolsep}{5pt}
\small

\begin{tabular}{lccc}
\toprule
\multicolumn{4}{c}{\textbf{PROBE}} \\
\midrule
\textbf{Target $\backslash$ Source} & origSCDM & noscrewSCDM & fixtureSCDM \\
\midrule
origSCDM      & \textbf{42.9} & 29.8 & 18.4 \\
noscrewSCDM   & 16.6 & \textbf{37.3} &  8.3 \\
fixtureSCDM   &  0.8 &  2.4 & \textbf{15.2} \\
\midrule
    \multicolumn{4}{c}{\textbf{Random Search}} \\
\midrule
\textbf{Target $\backslash$ Source} & origSCDM & noscrewSCDM & fixtureSCDM \\
\midrule
origSCDM      & \textbf{24.3} & 16.0 & 13.2 \\
noscrewSCDM   &  6.9 & \textbf{11.8} &  4.4 \\
fixtureSCDM   &  0.1 &  0.2 & \textbf{0.7} \\
\bottomrule
\end{tabular}
\end{table}

%% file: rq3_confidence.tex
\begin{table}[t]
\centering
\caption{Confidence and localization stability under non-failing perturbations. Values are reported as count (percentage of non-failing samples). Confidence and localization deviations are categorized as: \textit{minor} ($\leq 0.01$), \textit{small} ($0.01$--$0.05$], \textit{moderate} ($0.05$--$0.10$], and \textit{large} ($> 0.10$) (RQ3).}
\label{tab:rq3_stability}
\renewcommand{\arraystretch}{1.2}
\begin{tabular}{lrrr}
\toprule
\textbf{} & \textbf{origSCDM} & \textbf{noscrewSCDM} & \textbf{fixtureSCDM} \\
\midrule

\multicolumn{4}{l}{\textbf{Perturbations}} \\
Total 
& 132{,}400 & 160{,}000 & 28{,}400 \\
Non-failing 
& 97{,}520 (73.7\%) & 122{,}712 (76.7\%) & 25{,}459 (89.6\%) \\

\midrule
\multicolumn{4}{l}{\textbf{Confidence Stability}} \\
\textit{Minor} 
& 33{,}999 (34.9\%) & 52{,}385 (42.7\%) & 23{,}158 (91.0\%) \\
\textit{Small} 
& 46{,}188 (47.4\%) & 58{,}326 (47.5\%) & 2{,}067 (8.1\%) \\
\textit{Moderate} 
& 10{,}216 (10.5\%) & 8{,}774 (7.2\%) & 228 (0.9\%) \\
\textit{Large} 
& 7{,}117 (7.3\%) & 3{,}227 (2.6\%) & 6 (0.02\%) \\

\midrule
\multicolumn{4}{l}{\textbf{Localization Stability}} \\
\textit{Minor} 
& 28{,}902 (29.6\%) & 30{,}710 (25.0\%) & 19{,}642 (77.2\%) \\
\textit{Small} 
& 58{,}406 (59.9\%) & 70{,}474 (57.4\%) & 5{,}811 (22.8\%) \\
\textit{Moderate} 
& 10{,}118 (10.4\%) & 18{,}317 (14.9\%) & 6 (0.02\%) \\
\textit{Large} 
& 94 (0.1\%) & 3{,}211 (2.6\%) & 0 (0.0\%) \\

\bottomrule
\end{tabular}
\end{table}